\documentclass[runningheads]{llncs}
\usepackage{graphicx}
\usepackage{amsmath}
\usepackage{multirow}
\usepackage{amssymb}
\usepackage{hyperref}
\usepackage[misc]{ifsym}
\usepackage{colortbl}

\begin{document}
\title{SLPT: Selective Labeling Meets Prompt Tuning on Label-Limited Lesion Segmentation
}
\titlerunning{SLPT: Selective Labeling Meets Prompt Tuning}
%


\author{Fan Bai\inst{1,2,3} \and
Ke Yan\inst{2,3} \and
Xiaoyu Bai\inst{2,3} \and
Xinyu Mao\inst{1} \and
Xiaoli Yin\inst{4} \and
Jingren Zhou\inst{2,3} \and
Yu Shi\inst{4} \and
Le Lu\inst{2} \and
Max Q.-H. Meng\inst{1,5}}
%
\authorrunning{Fan Bai, et al.}
%

\institute{Department of Electronic Engineering, The Chinese University of Hong Kong,\\ Shatin, Hong Kong, China \\
\and
DAMO Academy, Alibaba Group \\
\and
Hupan Lab, 310023, Hangzhou, China \\
\and
Department of Radiology, Shengjing Hospital of China Medical University, Shenyang, 110004, China \\
\and
Department of Electronic and Electrical Engineering, Southern University of Science and Technology, Shenzhen, China\\
}

\maketitle              
\begin{abstract}
Medical image analysis using deep learning is often challenged by limited labeled data and high annotation costs. Fine-tuning the entire network in label-limited scenarios can lead to overfitting and suboptimal performance. Recently, prompt tuning has emerged as a more promising technique that introduces a few additional tunable parameters as prompts to a task-agnostic pre-trained model, and updates only these parameters using supervision from limited labeled data while keeping the pre-trained model unchanged. However, previous work has overlooked the importance of selective labeling in downstream tasks, which aims to select the most valuable downstream samples for annotation to achieve the best performance with minimum annotation cost. To address this, we propose a framework that combines selective labeling with prompt tuning (SLPT) to boost performance in limited labels. Specifically, we introduce a feature-aware prompt updater to guide prompt tuning and a TandEm Selective LAbeling (TESLA) strategy. TESLA includes unsupervised diversity selection and supervised selection using prompt-based uncertainty. In addition, we propose a diversified visual prompt tuning strategy to provide multi-prompt-based discrepant predictions for TESLA. We evaluate our method on liver tumor segmentation and achieve state-of-the-art performance, outperforming traditional fine-tuning with only 6\% of tunable parameters, also achieving 94\% of full-data performance by labeling only 5\% of the data.

\keywords{Active Learning \and Prompt Tuning \and Segmentation.}
\end{abstract}
\section{Introduction}
Deep learning has achieved promising performance in computer-aided diagnosis \cite{kim2019deep,isensee2021nnu,Bilic2019LiTS,simpson2019large}, but it relies on large-scale labeled data to train, which is challenging in medical imaging due to label scarcity and high annotation cost \cite{tajbakhsh2020embracing,bai2022discrepancy}. Specifically, expert annotations are required for medical data, which can be costly and time-consuming, especially in tasks such as 3D image segmentation. 

Transferring pre-trained models to downstream tasks is an effective solution for addressing the label-limited problem \cite{cheplygina2019not}, but fine-tuning the full network with small downstream data is prone to overfitting \cite{kumar2022fine}. Recently, prompt tuning \cite{liu2023pre,brown2020language} is emerging from natural language processing (NLP), which introduces additional tunable prompt parameters to the pre-trained model and updates only prompt parameters using supervision signals obtained from a few downstream training samples while keeping the entire pre-trained unchanged. By tuning only a few parameters, prompt tuning makes better use of pre-trained knowledge. It avoids driving the entire model with few downstream data, which enables it to outperform traditional fine-tuning in limited labeled data. Building on the recent success of prompt tuning in NLP \cite{brown2020language}, instead of designing text prompts and Transformer models, we explore visual prompts on Convolutional Neural Networks (CNNs) and the potential to address data limitations in medical imaging.

However, previous prompt tuning research \cite{liu2023pre,zhao2021prompt}, whether on language or visual models, has focused solely on the model-centric approach. For instance, CoOp \cite{zhou2022learning} models a prompt's context using a set of learnable vectors and optimizes it on a few downstream data, without discussing what kind of samples are more suitable for learning prompts. VPT \cite{jia2022visual} explores prompt tuning with a vision Transformer, and SPM \cite{liu2022prompt} attempts to handle downstream segmentation tasks through prompt tuning on CNNs, which are also model-centric. However, in downstream tasks with limited labeled data, selective labeling as a data-centric method is crucial for determining which samples are valuable for learning, similar to Active Learning (AL)\cite{settles2009active}. In AL, given the initial labeled data, the model actively selects a subset of valuable samples for labeling and improves performance with minimum annotation effort. Nevertheless, directly combining prompt tuning with AL presents several problems. First, unlike the task-specific models trained with initial data in AL, the task-agnostic pre-trained model (e.g., trained by related but not identical supervised or self-supervised task) is employed for data selection with prompt tuning. Second, in prompt tuning, the pre-trained model is frozen, which may render some AL methods inapplicable, such as those previously based on backbone gradient \cite{dai2020ggs} and feature \cite{parvaneh2022active}. Third, merging prompt tuning with AL takes work. Their interplay must be considered. However, previous AL methods \cite{zhan2022comparative} did not consider the existence of prompts or use prompts to estimate sample value.

Therefore, this paper proposes the first framework for selective labeling and prompt tuning (SLPT), combining model-centric and data-centric methods to improve performance in medical label-limited scenarios. We make three main contributions: (1) We design a novel feature-aware prompt updater embedded in the pre-trained model to guide prompt tuning in deep layers. (2) We propose a diversified visual prompt tuning mechanism that provides multi-prompt-based discrepant predictions for selective labeling. (3) We introduce the TESLA strategy which includes both unsupervised diversity selection via task-agnostic features and supervised selection considering prompt-based uncertainty. The results show that SLPT outperforms fine-tuning with just 6\% of tunable parameters and achieves 94\% of full-data performance by selecting only 5\% of labeled data.

\begin{figure}[tp!]
\centering
\includegraphics[width=\textwidth]{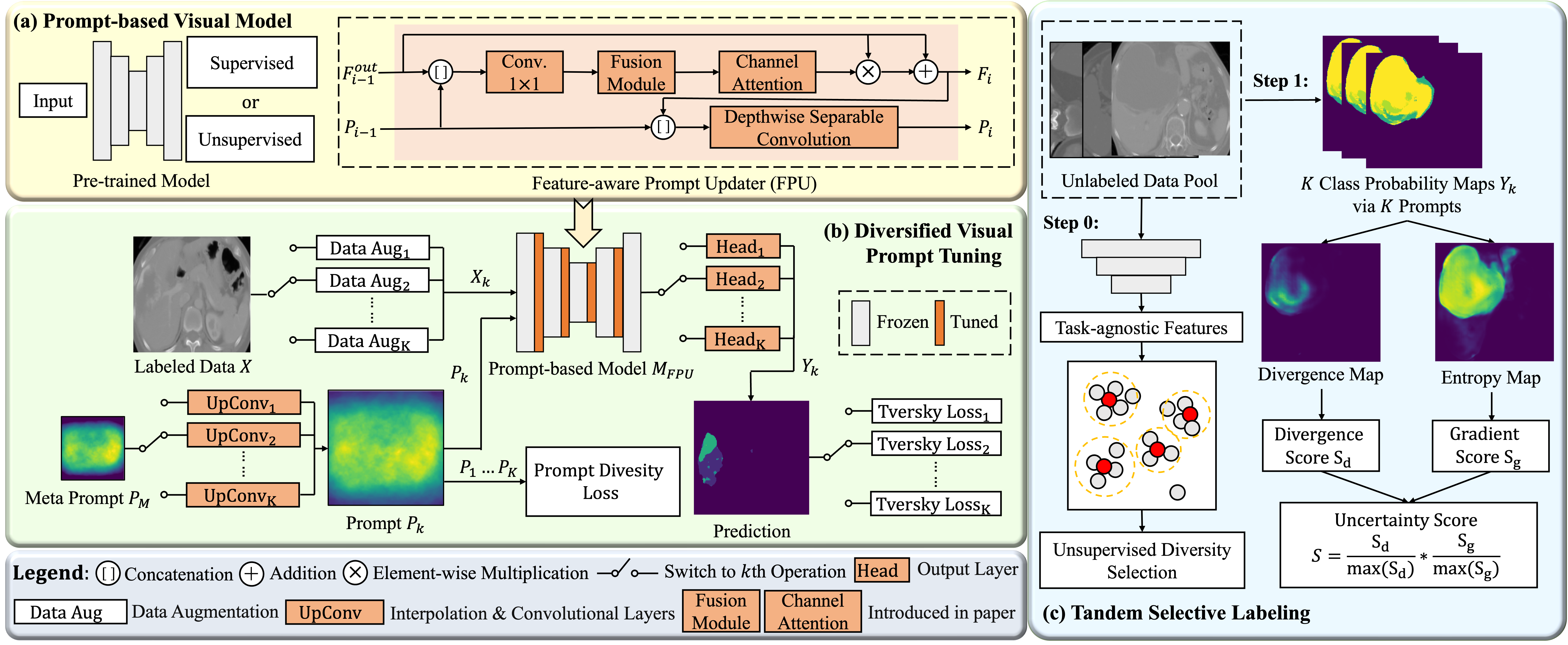}
\caption{Workflow of SLPT: \textbf{(1)} Create an initial label set via the pre-trained model for unsupervised diversity selection (subplot c step 0). \textbf{(2)} Insert a feature-aware prompt updater (subplot a) into the pre-trained model for prompt tuning with initial labels. \textbf{(3)} Use diversified visual prompt tuning (subplot b) to obtain prompt-based discrepant predictions. \textbf{(4)} Select valuable data by prompt-based uncertainty (subplot c step 1) and update the prompt-based model accordingly. Note: The orange modules are tunable for prompt tuning, while the gray ones are frozen. Please zoom in for details.} 
\label{fig1}
\end{figure}

\section{Methodology}
Given a task-agnostic pre-trained model and unlabeled data for an initial medical task, we propose SLPT to improve model performance. SLPT consists of three components, as illustrated in Fig.~\ref{fig1}: (a) a prompt-based visual model, (b) diversified visual prompt tuning, and (c) tandem selective labeling. Specifically, with SLPT, we can select valuable data to label and tune the model via prompts, which helps the model overcome label-limited medical scenarios.

\subsection{Prompt-based Visual Model}
The pre-trained model, learned by supervised or unsupervised training, is a powerful tool for improving performance on label-limited downstream tasks. Fine-tuning a large pre-trained model with limited data may be suboptimal and prone to overfitting \cite{kumar2022fine}. To overcome this issue, we draw inspiration from NLP \cite{liu2023pre} and explore prompt tuning on visual models. In order to facilitate prompt tuning on the model's deep layers, we introduce the Feature-aware Prompt Updater (FPU). FPUs are inserted into the network to update deep prompts and features. In Fig. \ref{fig1}(a), an FPU receives two inputs, feature map $F_{i-1}^{out}$ and prompt $P_{i-1}$, of the same shape, and updates to $F_i$ and $P_i$ through two parallel branches. In the feature branch, $F_{i-1}^{out}$ and $P_{i-1}$ are concatenated and fed into a 1x1 convolution and fusion module. The fusion module utilizes ASPP \cite{chen2017rethinking} to extract multi-scale contexts. Then a SE \cite{hu2018squeeze} module for channel attention enhances context by channel. Finally, the attention output and $F_{i-1}^{out}$ are element-wise multiplied and added to obtain the updated feature $F_i$. In the prompt branch, the updated feature $F_i$ is concatenated with the previous prompt $P_{i-1}$, and a parameter-efficient depth-separable convolution is employed to generate the updated prompt $P_i$.

To incorporate FPU into a pre-trained model, we consider the model comprising $N$ modular $M_i$ ($i=1,...,N$) and a head output layer. After each $M_i$, we insert an $FPU_i$. Given the input $F_{i-1}^{in}$ and prompt $P_{i-1}$, we have the output feature $F_{i}$, updated prompt $P_{i}$ and prediction $Y$ as follows:
\begin{equation}
F_{i-1}^{out} = M_i(F_{i-1}^{in}), \quad
F_{i}, P_{i} = \text{FPU}_i(F_{i-1}^{out}, P_{i-1}), \quad
Y = \text{Head}(F_{N})
\end{equation}
where input $X=F_0$, FPU and Head are tuned while $M_i$ is not tunable.

\subsection{Diversified Visual Prompt Tuning}
Inspired by multi-prompt learning \cite{liu2023pre} in NLP, we investigate using multiple visual prompts to evaluate prompt-based uncertainty. However, initializing and optimizing $K$ prompts directly can significantly increase parameters and may not ensure prompt diversity. To address these challenges, we propose a diversified visual prompt tuning approach. As shown in Fig. \ref{fig1}(b), our method generates $K$ prompts $P_k\in \mathbb{R}^{1\times D\times H\times W}$ from a meta prompt $P_M\in \mathbb{R}^{1\times \frac{D}{2}\times \frac{H}{2}\times \frac{W}{2}}$ through $K$ different upsampling and convolution operations $UpConv_k$. $P_M$ is initialized from the statistical probability map of the foreground category, similar to \cite{liu2022prompt}. Specifically, we set the foreground to 1 and the background to 0 in the ground-truth mask, and then average all masks and downsample to $1\times \frac{D}{2}\times \frac{H}{2}\times \frac{W}{2}$. To enhance prompt diversity, we introduce a prompt diversity loss $L_{div}$ that regularizes the cosine similarity between the generated prompts and maximizes their diversity. This loss is formulated as follows:
\begin{equation}
L_{div} = \sum_{k_1=1}^{K-1}\sum_{k_2=k_1+1}^{K}\frac{P_{k_1}\cdot P_{k_2}}{||P_{k_1}||_2 \cdot ||P{k_2}||_2}
\end{equation}
where $P_{k_1}$ and $P_{k_2}$ represent the $k_1$-th and $k_2$-th generated prompts, respectively, and $||\cdot||_{2}$ denotes the L2 norm. By incorporating the prompt diversity loss, we aim to generate a set of diverse prompts for our visual model.

In NLP, using multiple prompts can produce discrepant predictions \cite{allingham2023simple} that help estimate prompt-based uncertainty. Drawing inspiration, we propose a visual prompt tuning approach that associates diverse prompts with discrepant predictions. To achieve this, we design $K$ different data augmentation, heads, and losses based on corresponding $K$ prompts. By varying hyperparameters, we can achieve different data augmentation strengths, increasing the model's diversity and generalization. Different predictions $Y_k$ are generated by $K$ heads, each supervised with a Tversky loss \cite{salehi2017tversky} $TL_k = \frac{TP}{TP+\alpha_k FP+\beta_k FN}$, where TP, FP, and FN represent true positive, false positive, and false negative, respectively. To obtain diverse predictions with false positives and negatives, we use different $\alpha_k$ and $\beta_k$ values in $TL_k$. The process is formulated as follows:
\begin{equation}
    P_k = UpConv_k(P_M), \quad X_k = DA_k(X), \quad Y_k = Head_k(M_{FPU}(X_k,P_k))
\end{equation}
\begin{equation}
    L = \sum_{k=1}^{K}(\lambda_1 \cdot TL_k(Y_k,\hat{Y}) + \lambda_2 \cdot CE(Y_k, \hat{Y})) + \lambda_3 \cdot L_{div}
\end{equation}
where $k=1,...,K$, $M_{FPU}$ is the pre-trained model with FPU, $CE$ is the cross-entropy loss, and $\lambda_1=\lambda_2=\lambda_3=1$ weight each loss component. $\hat{Y}$ represents the ground truth and $L$ is the total loss.

\subsection{Tandem Selective Labeling}
Previous studies overlook the critical issue of data selection for downstream tasks, especially when available labels are limited. To address this challenge, we propose a novel strategy called TESLA. TESLA consists of two tandem steps: unsupervised diversity selection and supervised uncertainty selection. The first step aims to maximize the diversity of the selected data, while the second step aims to select the most uncertain samples based on diverse prompts.

\subsubsection{Step 0: Unsupervised Diversity Selection}
Since we do not have any labels in the initial and our pre-trained model is task-agnostic, we select diverse samples to cover the entire dataset. To achieve this, we leverage the pre-trained model to obtain feature representations for all unlabeled data. Although these features are task-independent, they capture the underlying relationships, with similar samples having closer feature distances. We apply the k-center method from Coreset \cite{sener2017active}, which identifies the $B$ samples that best represent the diversity of the data based on these features. These selected samples are then annotated and serve as the initial dataset for downstream tasks.

\subsubsection{Step 1: Supervised Uncertainty Selection}
After prompt tuning with the initial dataset, we obtain a task-specific model that can be used to evaluate data value under supervised training. Since only prompt-related parameters can be tuned while others are frozen, we assess prompt-based uncertainty via diverse prompts, considering inter-prompts uncertainty and intra-prompts uncertainty. In the former, we compute the multi-prompt-based divergence map $D$, given $K$ probability predictions $Y_k$ through $K$ diverse prompts $P_k$, as follows:
\begin{equation}
D = \sum_{k=1}^{K} \text{KL}(Y_k||Y_{\text{mean}}), \quad Y_{\text{mean}} = \frac{1}{K}\sum_{k=1}^{K}Y_k
\end{equation}
where $\text{KL}$ refers to the KL divergence\cite{kullback1951information}. Then, we have the divergence score $S_d = \text{Mean}(D)$, which reflects inter-prompts uncertainty.

In the latter, we evaluate intra-prompts uncertainty by computing the mean prediction of the prompts and propose to estimate prompt-based gradients as the model's performance depends on the update of prompt parameters $\theta_p$. However, for these unlabeled samples, computing their supervised loss and gradient directly is not feasible. Therefore, we use the entropy of the model's predictions as a proxy for loss. Specifically, we calculate the entropy-based prompt gradient score $S_g$ for each unlabeled sample as follows:
\begin{equation}
S_g = \sum_{\theta_p}||\nabla_{\theta_p}(-\sum Y_{mean} * \log Y_{mean})||_2
\end{equation}

To avoid manual weight adjustment, we employ multiplication instead of addition. We calculate our uncertainty score $S$ as follows:
\begin{equation}
S = \frac{S_d}{\max(S_d)} \times \frac{S_g}{\max(S_g)}
\end{equation}
where $\max(\cdot)$ finds the maximum value. We sort the unlabeled data by their corresponding $S$ values in ascending order and select the top $B$ data to annotate.

\section{Experiments and Results}
\subsection{Experimental Settings}
\subsubsection{Datasets and Pre-trained Model}
We conducted experiments on automating liver tumor segmentation in contrast-enhanced CT scans, a crucial task in liver cancer diagnosis and surgical planning~\cite{Bilic2019LiTS}. Although there are publicly available liver tumor datasets~\cite{Bilic2019LiTS,simpson2019large}, they only contain major tumor types and differ in image characteristics and label distribution from our hospital's data. Deploying a model trained from public data to our hospital directly will be problematic. Collecting large-scale data from our hospital and training a new model will be expensive. Therefore, we can use the model trained from them as a starting point and use SLPT to adapt it to our hospital with minimum cost. We collected a dataset from our in-house hospital comprising 941 CT scans with eight categories: hepatocellular carcinoma, cholangioma, metastasis, hepatoblastoma, hemangioma, focal nodular hyperplasia, cyst, and others. It covers both major and rare tumor types. Our objective is to segment all types of lesions accurately. We utilized a pre-trained model for liver segmentation using supervised learning on two public datasets \cite{simpson2019large} with no data overlap with our downstream task. The nnUNet\cite{isensee2021nnu} was used to preprocess and sample the data into 24x256x256 patches for training. To evaluate the performance, we employed a 5-fold cross-validation (752 for selection, 189 for test).

\subsubsection{Metrics}
We evaluated lesion segmentation performance using pixel-wise and lesion-wise metrics. For pixel-wise evaluation, we used the Dice per case, a commonly used metric \cite{Bilic2019LiTS}. For lesion-wise evaluation, we first do connected component analysis to predicted and ground truth masks to extract lesion instances, and then compute precision and recall per case \cite{powers2020evaluation}. A predicted lesion is regarded as a TP if its overlap with ground truth is higher than 0.2 in Dice.

\subsubsection{Competing Approaches}
In the prompt tuning experiment, we compared our method with three types of tuning: full parameter update (Fine-tuning, Learn-from-Scratch), partial parameter update (Head-tuning, Encoder-tuning, Decoder-tuning), and prompt update (SPM \cite{liu2022prompt}). 
In the unsupervised diversity selection experiment, we compared our method with random sampling.
In the supervised uncertainty selection experiment, we compared our method with random sampling, diversity sampling (Coreset \cite{sener2017active}, CoreCGN \cite{caramalau2021sequential}), and uncertainty sampling (Entropy, MC Dropout \cite{gal2016dropout}, Ensemble \cite{beluch2018power}, UncertainGCN \cite{caramalau2021sequential}, Ent-gn \cite{wang2022boosting}). Unlike Ensemble, our method was on multi-prompt-based heads. Furthermore, unlike Ent-gn, which computed the entropy-based gradient from a single prediction, we calculated a stable entropy from the muti-prompt-based mean predictions and solely considered the prompt gradient.

\subsubsection{Training Setup}
We conducted the experiments using the Pytorch framework on a single NVIDIA Tesla V100 GPU. The nnUNet \cite{isensee2021nnu} framework was used for 3D lesion segmentation with training 500 epochs at an initial learning rate of 0.01. We integrated 13 FPUs behind each upsampling or downsampling of nnUNet, adding only 2.7M parameters. During training, we set k=3 and employed diverse data augmentation techniques such as scale, elastic, rotation, and mirror. Three sets of TL parameters is ($\alpha_{1,2,3}$=0.5,0.7,0.3, $\beta_{1,2,3}$=0.5,0.3,0.7). To ensure fairness and eliminate model ensemble effects, we only used the model's prediction with $k=1$ during testing. We used fixed random seeds and 5-fold cross-validation for all segmentation experiments.

\begin{table}[tp!]
\centering
\caption{Evaluation of different tunings on the lesion segmentation with limited data (40 class-balanced patients). Prec. and Rec. denote precision and recall.}
\label{tab1}
\begin{tabular}{c|c|c|cll|cc|c} 
\hline
\multirow{2}{*}{Method} & \multirow{2}{*}{\begin{tabular}[c]{@{}c@{}}Tuning\\Type\end{tabular}} & \multirow{2}{*}{\begin{tabular}[c]{@{}c@{}}Trainable\\Parameters\end{tabular}} & \multicolumn{3}{c|}{Pixel-wise}                  & \multicolumn{2}{c|}{Lesion-wise} & \multirow{2}{*}{Mean}  \\ 
\cline{4-8}
                        &                                                                       &                                                                                & Dice           & Prec.          & Rec.           & Prec.          & Rec.            &                        \\ 
\hline
Fine-tuning             & \multirow{2}{*}{All}                                                  & 44.81M                                                                         & 64.43          & \textbf{87.69} & 59.86          & 50.84          & 54.14           & 63.39                  \\
Learn-from-Scratch      &                                                                       & 44.81M                                                                         & 54.15          & 73.33          & 50.25          & 45.84          & 45.78           & 53.87                  \\ 
\hline
Encoder-tuning          & \multirow{3}{*}{Part}                                                 & 19.48M                                                                         & 65.61          & 82.00          & 61.96          & 29.36          & 41.10           & 56.00                  \\
Decoder-tuning          &                                                                       & 23.64M                                                                         & 67.87          & 77.96          & \textbf{70.56} & 30.82          & 35.92           & 56.63                  \\
Head-tuning             &                                                                       & 0.10M                                                                          & 56.73          & 74.45          & 55.57          & 23.29          & 29.74           & 47.96                  \\ 
\hline
SPM \cite{liu2022prompt}                    & \multirow{2}{*}{Prompt}                                               & 3.15M                                                                          & 68.60          & 83.07          & 69.02          & 62.15          & 55.19           & 67.61                  \\
Ours                    &                                                                       & \textbf{2.71M}                                                                 & \textbf{68.76} & 79.63          & 69.76          & \textbf{64.63} & \textbf{61.18}  & \textbf{68.79}         \\
\hline
\end{tabular}
\end{table}

\subsection{Results}
\subsubsection{Evaluation of Prompt Tuning}
Since we aim to evaluate the efficacy of prompt tuning on limited labeled data in Table~\ref{tab1}, we create a sub-dataset of approximately 5\% (40/752) from the original dataset. Specifically, we calculate the class probability distribution vector for each sample based on the pixel class in the mask and use CoreSet with these vectors to select 40 class-balanced samples. Using this sub-dataset, we evaluated various tuning methods for limited medical lesion diagnosis data. The results are summarized in Table~\ref{tab1}. Fine-tuning all parameters served as the strongest baseline, but our method, which utilizes only 6\% tunable parameters, outperformed it by 5.4\%. Although SPM also outperforms fine-tuning, our methods outperform SPM by 1.18\% and save 0.44M tunable parameters with more efficient FPU. In cases of limited data, fine-tuning tends to overfit on a larger number of parameters, while prompt tuning does not. The pre-trained model is crucial for downstream tasks with limited data, as it improves performance by 9.52\% compared to Learn-from-Scratch. Among the three partial tuning methods, the number of tuning parameters positively correlates with the model's performance, but they are challenging to surpass fine-tuning.

\begin{table}[tp!]
\centering
\caption{Comparison of data selection methods for label-limited lesion segmentation. Step 0: unsupervised diversity selection. Step 1: supervised uncertainty selection. The labeling budget for each step is 20 patients. Step +$\infty$ refers to fully labeled 752 data.}
\label{tab2}
\arrayrulecolor{black}
\setlength\tabcolsep{6pt}
\begin{tabular}{c|c|cll|cc|c} 
\hline
\multirow{2}{*}{Step}                         & \multirow{2}{*}{Method}                                                     & \multicolumn{3}{c|}{Pixel-wise}                       & \multicolumn{2}{c|}{Lesion-wise}     & \multirow{2}{*}{Mean}  \\ 
\cline{3-7}
                                              &                                                                             & Dice           & Prec.          & Rec.           & Prec.          & Rec.           &                        \\ 
\hline
\multirow{2}{*}{0}                            & Random                                                                      & 65.58          & 80.00          & 65.21          & 23.46          & 39.94          & 54.84                  \\
                                              & Ours                                                                        & 68.20          & 78.97          & 69.15          & 32.51          & 34.67          & 56.70                  \\ 
\hline
\multirow{11}{*}{1}                           & Random                                                                      & 66.67          & 79.95          & 70.67          & 41.45          & 39.45          & 59.64                  \\
                                              & Entropy                                                                     & 66.39          & 80.85          & 66.96          & 37.40          & 39.47          & 58.21                  \\
                                              & MC Dropout \cite{gal2016dropout}                                                                  & 69.23          & 79.61          & 69.48          & 30.43          & 36.29          & 57.01                  \\
                                              & Ensemble \cite{beluch2018power}                                                                 & 69.79          & 80.25          & 69.54          & \textbf{64.38} & 58.34          & 68.46                  \\
                                              & CoreSet \cite{sener2017active}                                                                    & 70.72          & 79.34          & 72.03          & 46.03          & 51.24          & 63.87                  \\
                                              & CoreGCN \cite{caramalau2021sequential}                                                                  & 70.91          & 77.56          & 72.37          & 51.73          & 49.88          & 64.49                  \\
                                              & UncertainGCN \cite{caramalau2021sequential}                                                                  & 71.44          & 75.07          & \textbf{75.62} & 72.83          & 44.99          & 67.99                  \\
                                              & Ent-gn \cite{wang2022boosting}                                                                     & 70.54          & 79.91          & 71.42          & 61.12          & 56.37          & 67.87                  \\ 
\cline{2-8}
                                              & Ours (w/o $S_d$)                                                         & 69.54          & 81.97          & 68.59          & 60.47          & 59.82          & 68.08                  \\
                                              & Ours (w/o $S_g$)                                                         & 71.01          & 80.68          & 69.83          & 59.42          & 58.78          & 67.94                  \\
                                              & Ours                                                                        & \textbf{72.07} & \textbf{82.07} & 72.37          & 61.21          & \textbf{61.90} & \textbf{69.92}         \\ 
\hline
\textcolor[rgb]{0.302,0.318,0.337}{+$\infty$} & \begin{tabular}[c]{@{}c@{}}Fine-tuning with\\Full Labeled Data\end{tabular} & 77.44          & 85.44          & 77.15          & 62.78          & 68.56          & 74.27                  \\
\hline
\end{tabular}
\end{table}

\subsubsection{Evaluation of Selective Labeling}
We conducted steps 0 (unsupervised selection) and 1 (supervised selection) from the unlabeled 752 data and compared our approach with other competing methods, as shown in Table~\ref{tab2}. In step 0, without any labeled data, our diversity selection outperformed the random baseline by 1.86\%. Building upon the 20 data points selected by our method in step 0, we proceeded to step 1, where we compared our method with eight other data selection strategies in supervised mode. As a result, our approach outperformed other methods because of prompt-based uncertainty, such as Ent-gn and Ensemble, by 2.05\% and 1.46\%, respectively. Our approach outperformed Coreset by 6.05\% and CoreGCN by 5.43\%. We also outperformed UncertainGCN by 1.93\%. MC Dropout and Entropy underperformed in our prompt tuning, likely due to the difficulty of learning such uncertain data with only a few prompt parameters. Notably, our method outperformed random sampling by 10.28\%. These results demonstrate the effectiveness of our data selection approach in practical tasks. 

\subsubsection{Ablation Studies}
We conducted ablation studies on $S_d$ and $S_g$ in TESLA. As shown in Table~\ref{tab2}, the complete TESLA achieved the best performance, outperforming the version without $S_d$ by 1.84\% and the version without $S_g$ by 1.98\%. It shows that each component plays a critical role in improving performance.

\section{Conclusions}
We proposed a pipeline called SLPT that enhances model performance in label-limited scenarios. With only 6\% of tunable prompt parameters, SLPT outperforms fine-tuning due to the feature-aware prompt updater. Moreover, we presented a diversified visual prompt tuning and a TESLA strategy that combines unsupervised and supervised selection to build annotated datasets for downstream tasks.
SLPT pipeline is a promising solution for practical medical tasks with limited data, providing good performance, few tunable parameters, and low labeling costs. Future work can explore the potential of SLPT in other domains.

\subsubsection{Acknowledgements.} The work was supported by Alibaba Research Intern Program. Fan Bai and Max Q.-H. Meng were supported by National Key R\&D program of China with Grant No. 2019YFB1312400, Hong Kong RGC CRF grant C4063-18G, and Hong Kong Health and Medical Research Fund (HMRF) under Grant 06171066. Xiaoli Yin and Yu Shi were supported by National Natural Science Foundation of China (82071885).

%
%
%
\bibliographystyle{splncs04}
\bibliography{miccai}

\end{document}